\documentclass{article}

\PassOptionsToPackage{numbers}{natbib}  

\usepackage[preprint]{neurips_2026}

\usepackage{graphicx}
\usepackage{amsmath}
\usepackage{algorithm}
\usepackage{algpseudocode}
\usepackage{wrapfig}

\usepackage[utf8]{inputenc} 
\usepackage[T1]{fontenc}    
\usepackage{hyperref}       
\hypersetup{hidelinks}      
\usepackage{url}            
\usepackage{booktabs}       
\usepackage{amsfonts}       
\usepackage{nicefrac}       
\usepackage{microtype}      
\usepackage{xcolor}         
\usepackage{amsmath}
\usepackage{amssymb}

\usepackage{graphicx}
\usepackage{multirow}
\usepackage{array}
\usepackage{enumitem}

\title{When Eyes Betray AI: Social Gaze Consistency as a Semantic Cue for AI-Generated Image Detection}

\author{%
  Jihyeon Kim \\
  School of Computer Engineering \\
  Hoseo University \\
  \texttt{20220991@vision.hoseo.edu} \\
  \And
  Sohee Kim \\
  School of Electronic Engineering \\
  Soongsil University \\
  \texttt{soheekim@soongsil.ac.kr} \\
  \And
  Soosan Lee \\
  School of Electronic Engineering \\
  Soongsil University \\
  \texttt{susana0221@soongsil.ac.kr} \\
  \And
  Souhwan Jung \\
  School of Electronic Engineering \\
  Soongsil University \\
  \texttt{souhwanj@ssu.ac.kr} \\
  \And
  James Matthew Rehg \\
  School of Computer Science \\
  University of Illinois at Urbana-Champaign \\
  \texttt{jrehg@illinois.edu} \\
  \And
  Hyesong Choi \\
  School of Electronic Engineering \\
  Soongsil University \\
  \texttt{hyesong@ssu.ac.kr} \\
}

\begin{document}

\raggedbottom

\maketitle

\begin{abstract}
Recent generative models have largely closed the gap on low-level artifacts---pixel fingerprints, frequency anomalies, upsampling traces---particularly in person-centric and partial-edit settings where the manipulated region is small and surrounded by photometrically authentic content. We introduce Social Gaze Consistency, a high-level semantic cue defined as the mutual coherence of gaze direction, head--eye alignment, and pupil placement between interacting individuals, and show that it constitutes a previously underutilized detection axis orthogonal to existing low-level paradigms. We instantiate this insight through three coupled mechanisms: (i) a controlled diagnostic dataset with region-specific perturbations of gaze-consistent imagery, where strict pair-level grouping forecloses generator-fingerprint memorization as an optimization-time shortcut rather than relying on augmentation; (ii) Block-Compositional Caption Supervision, which holds a single 5-block reasoning skeleton invariant across $1{,}250$ macro-combined captions, decoupling reasoning consistency from surface diversity; (iii) Cross-architecture validation showing the same supervision improves a vision-language backbone (FakeVLM) by $+3.7$\,pp on the COCOAI Interaction subset (balanced accuracy $67.8 \to 71.5$) and $+1.3$\,pp on the COCOAI Person subset ($83.0 \to 84.3$), with consistent gains on a vision-only \mbox{backbone} (Effort), evidencing a backbone-agnostic cue. Real- and fake-class recalls rise simultaneously, ruling out a ``predict-all-fake'' artifact. A four-step mechanistic account---paired-edit shortcut blocking, hard-to-easy difficulty transfer, CLIP prior preservation, and diffusion-family shared spectral weakness in periocular structure---explains why training on a single inpainter (FLUX.1-Fill) transfers to multi-generator suites. We will release the code upon acceptance to facilitate reproducibility.
\end{abstract}

\section{Introduction}

The trajectory of modern generative models---Stable Diffusion 3~\citep{esser2024sd3}, FLUX.1, DALL$\cdot$E 3~\citep{betker2023dalle3}, Midjourney v6---has progressively dismantled the low-level signals upon which the dominant lineage of AI-generated content (AIGC) detectors~\citep{wang2020cnnspot, ojha2023univfd, tan2024npr, yan2025aide} was built. Frequency-domain anomalies~\citep{frank2020frequency}, neighboring-pixel relationships introduced by upsampling~\citep{durall2020upconv}, characteristic skin-texture smoothing, and finger-count irregularities have all been substantially mitigated by successive model releases. The empirical consequence is striking: detectors such as NPR~\citep{tan2024npr}, UnivFD~\citep{ojha2023univfd}, and AIDE~\citep{yan2025aide}, which posted near-saturated accuracy on prior benchmarks, collapse to near-chance levels on contemporary multi-generator suites that involve human subjects (Section~\ref{sec:baselines}). This is not a mere distributional shift; it is the partial exhaustion of a cue space.

Compounding this, the threat surface itself has broadened. Beyond fully synthesized imagery, \emph{partial-edit fakes}---in which a localized region of a real photograph is regenerated by an inpainting model~\citep{lugmayr2022repaint, nichol2022glide}---now constitute a practically significant attack vector spanning identity manipulation, social-media misinformation, and targeted facial-expression edits. Such fakes inherit the photometric statistics of authentic imagery everywhere except a small region, evading detectors that rely on globally distributed cues.

We argue that detection cues lie on at least two axes. The low-level axis comprises pixel statistics, frequency residuals, and architectural fingerprints---the territory progressively reclaimed by generators. The high-level semantic-consistency axis comprises socially meaningful relations between depicted entities~\citep{cao2025socialgesture}: the geometric coherence of mutual gaze, the alignment of head pose with eye direction, and the plausibility of attentional targets in multi-person scenes. The standard denoising diffusion objective does not explicitly enforce these higher-order relations: nothing penalizes a generated face whose eyes drift fractionally off the partner's gaze target while remaining locally photorealistic.

Earlier work in social-interaction recognition (notably LAEO-Net~\citep{marin2019laeonet}) established mutual gaze as a primary signal for video-based interaction parsing. We transpose this insight into a previously unexamined setting: single-image AIGC detection, where mutual gaze becomes a falsifiable geometric prediction rather than an action-recognition target.

Person-centric and interaction-centric content constitutes the weakest territory for the \emph{current generation} of detectors---precisely where a semantic cue should help the most. On the multi-person interaction subset we curate from MS COCOAI~\citep{roy2026cocoai}, the dominant low-level detectors collapse to near-random performance(BA, \%): AIDE~\citep{yan2025aide} ($48.7$), NPR~\citep{tan2024npr} ($51.5$), and UnivFD~\citep{ojha2023univfd} ($50.0$). Even shortcut-robust detectors such as Effort~\citep{yan2025effort} ($59.0$) and SIDA-13B~\citep{huang2025sida} ($60.9$) remain stuck in the $50$--$60$ range, with the same pattern persisting on the broader person-centric subset---motivating an orthogonal, semantically grounded detection axis.

We make three primary contributions, supported by a \emph{Custom Gaze} diagnostic dataset constructed for these experiments.

\textbf{(C1) A new cue.} We introduce \emph{Social Gaze Consistency} as a semantic detection axis distinct from low-level paradigms and prior single-person periocular forensics, with cross-generator transferability substantiated by a four-step mechanism.

\textbf{(C2) A supervision design.} We propose \emph{Block-Compositional Caption Supervision}, whose macro-pool sampling produces $1{,}250$ unique captions with an invariant reasoning skeleton from only $20$ macro-pool entries.

\textbf{(C3) Cross-architecture validation.} The same supervision improves FakeVLM by $+3.7$\,pp on the COCOAI interaction-centric subset (BA $67.8 \to 71.5$) and $+1.3$\,pp on the person-centric subset ($83.0 \to 84.3$), with consistent gains on Effort, evidencing an anatomically grounded, generator-agnostic signal.

The supporting Custom Gaze dataset---$46{,}830$ paired real/fake images with region-specific perturbations of gaze-consistent imagery (Appendix~\ref{app:datasheet})---underpins all three contributions through pair-level identity preservation that forecloses generator-fingerprint memorization.

\section{Related Work}

\subsection{Vision-Language Detectors}

The emergence of large multimodal models (LMMs) such as LLaVA~\citep{liu2023llava, liu2024llava15} has enabled fake detectors that produce explanatory natural-language output rather than binary scores. The decisive empirical observation comes from FakeVLM~\citep{wen2025fakevlm}: fine-tuning LLaVA-v1.5 on FakeClue (an explanation corpus of $\approx\!104$K AI-generated images paired with free-form captions describing visible synthesis artifacts) and switching the output head from a linear classifier to free-form text yields a $+2.7$\,pp gain on LOKI~\citep{ye2025loki}---a controlled isolation of how supervision \emph{form} alone shapes detection performance. SIDA~\citep{huang2025sida} extends this paradigm with a 13B-parameter LISA backbone and adds tampered-region localization. This lineage situates explanatory generation as a \emph{learning signal} rather than post-hoc interpretation: by requiring the model to articulate \emph{why} an image is fake during training, optimization is forced to internalize discriminative features at the conceptual level rather than at pixel statistics.

Our work refines this lineage at a finer granularity. Where FakeVLM establishes that \emph{explanatory generation} (vs.\ binary classification) helps, we ask the next-level question: \emph{which structural form} of explanation maximizes detection signal. We isolate the effect of injecting a fixed reasoning skeleton into supervision, decoupling reasoning consistency from raw output diversity.

\subsection{Mutual Gaze in Recognition and Periocular Cues in Forensics}

\textbf{Mutual gaze in social-interaction recognition.} LAEO-Net~\citep{marin2019laeonet} introduced a three-branch CNN for ``people looking at each other'' detection in video, demonstrating that mutual gaze geometry is a high-information signal for social-interaction parsing. OI Mutual Gaze~\citep{doosti2021oimg} released the first large-scale image-level mutual-gaze dataset with binary annotations on 29.2K Open Images. A complementary research line on gaze following~\citep{recasens2015gazefollow} and attention target detection~\citep{chong2020atttargets} infers \emph{where} a person is looking from a single image or video, supplying the geometric primitive (head $\to$ attentional target) whose multi-person consistency our forensic cue exploits. Foundational gaze-estimation backbones~\citep{kellnhofer2019gaze360, zhang2020ethxgaze} further establish that head-pose and pupil-direction estimation have reached sufficient maturity for downstream forensic consumption.

\textbf{Eye and gaze cues in deepfake forensics.} A separate lineage exploits eye-region cues for deepfake detection: eye-blinking inconsistency~\citep{li2018ictuoculi} for first-generation GAN deepfakes, and pupil-shape irregularities~\citep{guo2022pupil} and iris-region forensics~\citep{hu2021corneal} for anatomically inconsistent ocular structure. These approaches all operate at the \emph{single-person} level, exploiting intra-face anatomical priors.

\textbf{Our positioning.} Unlike LAEO-Net and OI Mutual Gaze, which use mutual gaze as an action-recognition or social-interaction \emph{target}, we treat it as a \emph{falsifiable forensic prediction}---a relation that authentic photographs satisfy by construction and that contemporary inpainting models systematically violate at the eye region. Unlike single-person eye-cue detectors, which exploit \emph{intra-face} anatomical priors, our cue is fundamentally \emph{inter-person}: the geometric coherence of two interacting individuals' gaze vectors with their respective head poses.

\section{Method}
\label{sec:method}

\begin{figure}[!htb]
  \centering
  \includegraphics[width=\linewidth]{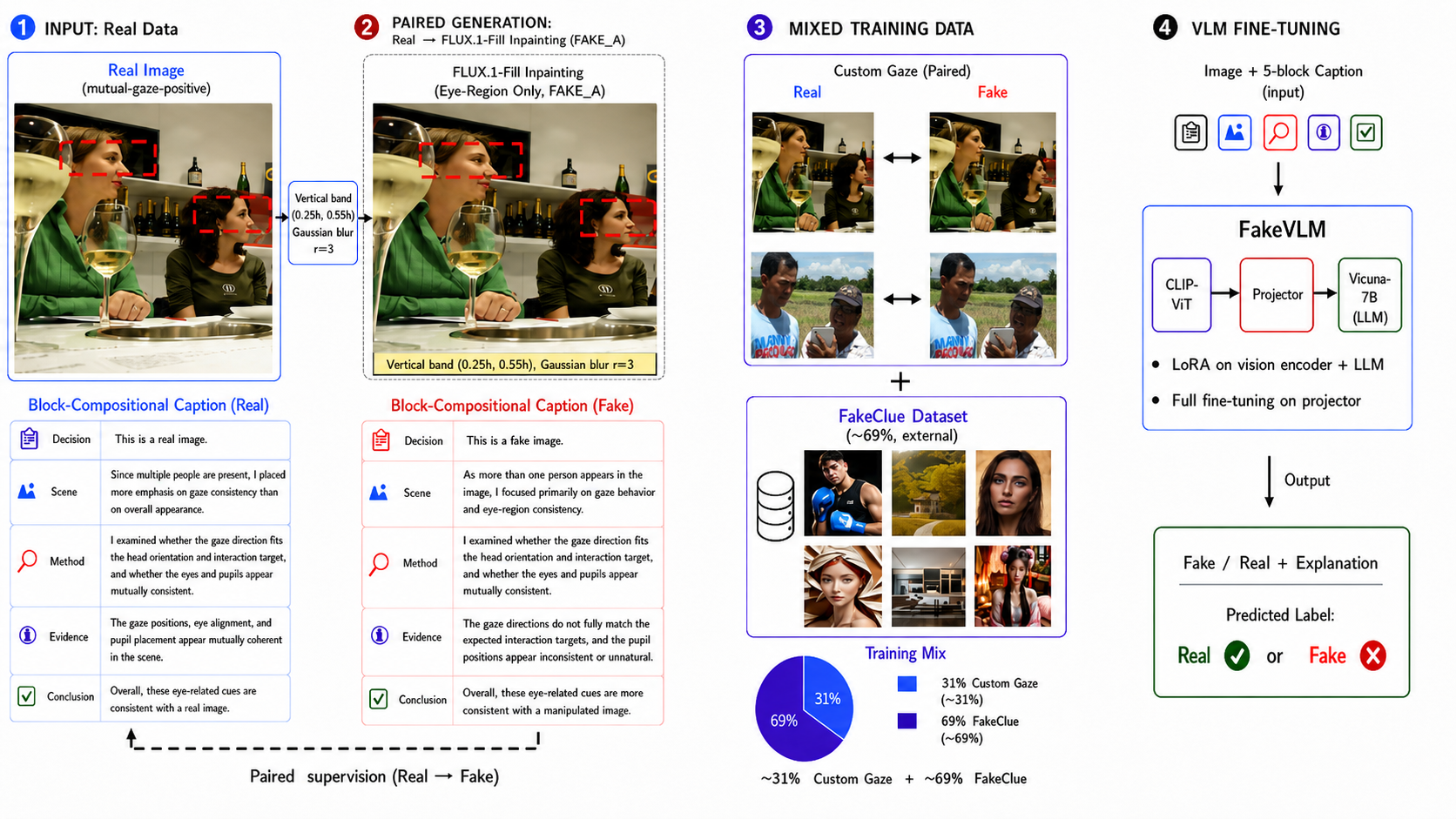}
  \caption{Custom Gaze construction and Block-Compositional Caption Supervision. Only the eye region is regenerated by FLUX.1-Fill; the paired real--fake samples share an invariant 5-block reasoning skeleton (Decision $\to$ Scene $\to$ Method $\to$ Evidence $\to$ Conclusion) with surface variation across $1{,}250$ macro-combined captions.}
  \label{fig:method_overview}
\end{figure}

We formalize a person-centric AIGC detection problem and present three coupled components (Figure~\ref{fig:method_overview}): (i) a paired-edit data construction that isolates the gaze cue from confounding generative fingerprints; (ii) a Block-Compositional Caption schema that injects a reasoning skeleton into supervision while preserving surface diversity; and (iii) a mixed fine-tuning protocol with balanced-accuracy checkpoint selection.

\subsection{Problem Formulation}

A detector outputs both a binary authenticity label ($1{=}$real, $0{=}$fake) and a natural-language explanation $c\in\mathcal{T}$; following FakeVLM, we exploit $\mathcal{T}$ as a supervision channel rather than as post-hoc interpretation.

We posit a two-axis cue decomposition: $\varphi_{\mathrm{low}}$ captures low-level statistics (frequency, pixel residuals, upsampling traces) and $\varphi_{\mathrm{high}}$ captures semantic-consistency descriptors (gaze geometry, head--eye alignment, attentional targets). Within the person-centric subdomain under partial-edit attacks, we put forward the working hypothesis
\begin{equation*}
\tag*{(H1)}
I(Y\,;\,\varphi_{\mathrm{high}}(x)) > I(Y\,;\,\varphi_{\mathrm{low}}(x)),
\end{equation*}
tested behaviorally via downstream accuracy on partial-edit and interaction-centric subsets.

\subsection{Custom Gaze Dataset Construction}
\label{sec:method-data}

Our dataset construction is governed by a single design principle: pair-level identity preservation. Each fake image shares base identity, scene, lighting, and global composition with a corresponding real image, differing only within a localized eye region. This structure enables shortcut blocking during learning (cf.\ \S\ref{sec:cross-arch}, M1): paired supervision forecloses memorizing FLUX-specific fingerprints.

\paragraph{Real source.} Our real source is the mutual-gaze-positive subset of OI Mutual Gaze~\citep{doosti2021oimg} ($23{,}415$ mutual-gaze-positive image-pairs derived from $8{,}972$ retained source images via multi-bbox unpacking; (details in Appendix~\ref{app:custom-gaze}); each retained image carries two face bounding boxes $(x_{\min}, y_{\min}, x_{\max}, y_{\max})$ for participants A and B.

\paragraph{Eye-region masking and inpainting.} Given a face bounding box, we compute the eye-region mask $M_{\mathrm{eye}}$ via
\begin{equation}
M_{\mathrm{eye}} = \{(i,j) : y_{\min} + 0.25h \le i \le y_{\min} + 0.55h 
\text{ and } x_{\min} + 0.05w \le j \le x_{\max} - 0.05w\}
\end{equation}
where $h \triangleq y_{\max} - y_{\min}$ and $w \triangleq x_{\max} - x_{\min}$. The vertical band $[0.25\,h,\, 0.55\,h]$ empirically isolates the eye region across pose variations while excluding nose, mouth, hair, and background, and a Gaussian blur of radius $3$ softens mask boundaries. We employ FLUX.1-Fill-dev~\citep{bfl2024flux} (a 12B rectified flow transformer; abbreviated FLUX.1-Fill in subsequent text) in bfloat16 at $20$ inference steps, guidance scale $25$ (see Appendix~\ref{app:custom-gaze} for full pipeline); images are resized to max side $1024$ px and aligned to a 16-pixel grid per FLUX requirement.

\paragraph{Final composition.} Custom Gaze comprises $46{,}830$ images split $8\!:\!1\!:\!1$ (train\,:\,val\,:\,test) with stratification and \emph{pair-level grouping} to prevent leakage; the test split is the partial-edit hard-fake benchmark used throughout. The datasheet is in Appendix~\ref{app:datasheet}.

\subsection{Block-Compositional Caption Supervision}
\label{sec:method-caption}

Recent advances in instruction-tuned LLMs~\citep{liu2023llava,wen2025fakevlm} have established that \emph{what the model is asked to output} is at least as consequential as \emph{what it is asked to predict}. Table~6 of Wen et al.~\citep{wen2025fakevlm} isolates this effect: switching the output head from a linear classifier to free-form text yields $+2.7$\,pp on LOKI. We hypothesize a finer claim: the \emph{structural depth} of the requested output, when properly compositionally encoded, induces detector behavior qualitatively different from either binary classification or unstructured generation.

We instantiate this through a 5-tuple caption schema $c = (p_1, p_2, p_3, p_4, p_5)$ corresponding to Decision (label header), Scene (multi-person context), Method (cue-type declaration: gaze-based reasoning), Evidence (gaze-geometric observation), and Conclusion (verdict synthesis). Position $p_1$ is deterministic given the label ($p_1(\text{real}) = $ ``This is a real image.''; $p_1(\text{fake}) = $ ``This is a fake image.''), while positions $p_2$ through $p_5$ are populated by sampling from per-position macro pools of cardinality $5$, yielding $|\mathcal{C}| = 1 \times 5^4 \times 2 = 1{,}250$ unique captions distributed approximately uniformly across $18{,}732$ training samples per label. The \emph{reasoning skeleton} (Scene $\rightarrow$ Method $\rightarrow$ Evidence $\rightarrow$ Conclusion) is invariant across all $1{,}250$ captions; only the surface lexicalization of each block varies.

Why compositional, not free-form? A natural alternative is to delegate caption generation to a stronger LMM (e.g., GPT-4o, Qwen2-VL-72B), as FakeVLM does for its FakeClue corpus. This works for general AIGC detection but not for gaze: in qualitative pilot studies, off-the-shelf LMMs defaulted to texture-based reasoning even when explicitly prompted toward gaze cues, failing to instill consistent gaze-centric reasoning. Our schema resolves this through a single reasoning path $\times$ 1,250 surface variants design choice, which simultaneously achieves (i) mode collapse prevention, with unique output ratio ranging $0.93$--$0.97$ across post-warmup 
training (mean $0.94$; $0.93$ at the deployed step $1{,}650$; 
$0.96$ over the last $50$ evaluations; full diagnostics in 
Appendix~\ref{app:mode-collapse-detail}); (ii) reasoning coherence, with $\textsc{top1\_template\_ratio}$ stabilizing at $0.52$--$0.53$; and (iii) annotation cost reduction from $\mathcal{O}(N)$ per-sample human authoring to a constant $20$ macro-pool entries ($4$ positions $\times$ $5$ variants), which yield tens of thousands of captions via combinatorial sampling.

\paragraph{Training objective.} The supervision channel is realized through standard token-level cross-entropy on the assistant's response tokens (with question tokens masked)
\begin{equation}
\mathcal{L}_{\mathrm{caption}}(\theta) = -\,\mathbb{E}_{(x,c) \sim \mathcal{D}} \left[ \sum_{t=1}^{|c|} \log p_\theta\!\left(c_t \;\middle|\; c_{<t},\, x \right) \right],
\end{equation}
with the question turn (``Does the image look real/fake?'') masked from the loss. At inference, we extract the label via a deterministic regex matching the prefix \texttt{This is a (real|fake) image.} over $p_1$, requiring no learned classification head.

\paragraph{Mix composition.} To prevent person-only training from inducing distributional collapse on non-person categories, we mix the Custom Gaze dataset ($46{,}830$ samples $=23{,}415$ paired images across the train/val/test splits, with the pair-level grouping of \S\ref{sec:method-data} ensuring that val/test pairs are excluded from gradient updates) with the $104{,}343$-sample FakeClue training set~\citep{wen2025fakevlm}, yielding $151{,}173$ total samples ($\approx 31\%$ gaze, $\approx 69\%$ FakeClue). FakeClue captions are retained in their original free-form form, providing complementary texture-centric supervision.

\subsection{Mixed Fine-Tuning with Balanced-Accuracy Checkpointing}
\label{sec:method-finetune}

\paragraph{Trainable parameters.} We adopt FakeVLM (LLaVA-v1.5 + Vicuna-v1.5-7B + CLIP-ViT-L/14-336) as our base and apply parameter-efficient fine-tuning differentially across three modules:

\begin{table}[h]
\centering
\small
\caption{Adaptation configuration: LoRA on vision encoder and LLM; 
full fine-tuning on the projector as the cross-modal bottleneck.}
\begin{tabular}{ll}
\toprule
\textbf{Module} & \textbf{Adaptation} \\
\midrule
Vision encoder (CLIP-ViT) & LoRA, all attention and MLP linear layers \\
Vision-language projector & Full fine-tuning (\texttt{modules\_to\_save}) \\
LLM (Vicuna-7B) & LoRA, all attention and MLP linear layers \\
\bottomrule
\end{tabular}
\end{table}

LoRA~\citep{hu2022lora} hyperparameters: rank $r{=}16$; scaling $\alpha{=}32$; dropout $0.05$; bias none. Trainable parameters constitute approximately $0.6\%$ of the base model ($\sim$45M of $\sim$7B). The asymmetry reflects that the projector is the \emph{modality bridge} where gaze-cue integration most directly requires reweighting of cross-modal correspondences.

\paragraph{Optimization.} We optimize with AdamW~\citep{loshchilov2019adamw} ($\beta = (0.9, 0.999)$, $\varepsilon = 10^{-8}$, $\lambda{=}0$) at peak learning rate $2 \times 10^{-5}$ under a cosine schedule with warmup ratio $0.03$. The effective batch size is $32$ (per-device $1$, gradient accumulation $4$, $8$ GPUs) on RTX A6000s with DeepSpeed ZeRO-2~\citep{rajbhandari2020zero} and bfloat16 precision. Training proceeds for $2$ epochs ($7{,}558$ steps), with sequence maximum length $1{,}024$ tokens and random seed fixed to $42$.

\paragraph{Balanced-accuracy checkpoint selection.} We select the deployed checkpoint by balanced accuracy rather than validation loss. With sensitivity $\mathrm{TPR} \triangleq \mathrm{TP}/(\mathrm{TP} + \mathrm{FN})$ and specificity $\mathrm{TNR} \triangleq \mathrm{TN}/(\mathrm{TN} + \mathrm{FP})$, balanced accuracy is $\mathrm{BA} = (\mathrm{TPR} + \mathrm{TNR})/2$~\citep{brodersen2010balanced}. For explanatory generative detectors, validation loss and BA decouple: a model can reduce token-level loss by drifting toward template uniformity while sacrificing class-balanced classification competence. Empirically, our loss minimum occurs at step $2{,}850$ (loss $= 0.2252$) while BA peaks at step $1{,}650$ ($\mathrm{BA} = 0.9990$)---a gap of $1{,}200$ steps. We therefore select
\begin{equation}
\theta^{*} = \arg\max_{\theta_t} \mathrm{BA}(\theta_t \,;\, \mathcal{D}_{\mathrm{dev}}),
\end{equation}
with $\mathcal{D}_{\mathrm{dev}}$ a held-out $970$-sample mixture ($470$ Custom Gaze + $500$ FakeClue test samples), evaluated every $50$ steps. We refer to the resulting checkpoint as \textsc{mix1650} and use it as the deployed model throughout the experiments.

\section{Experiments}
\label{sec:experiments}

\subsection{Experimental Protocol}

\paragraph{Evaluation suites.} We evaluate on three benchmarks chosen to triangulate distinct failure modes of multi-generator AIGC: Custom Gaze (partial-edit hard fakes), COCOAI Person (single-person AIGC across five generators)~\citep{roy2026cocoai}, and COCOAI Interaction~$\bigstar$ (multi-person interaction; our \emph{core evidence} benchmark)~\citep{roy2026cocoai}. Per-benchmark construction---caption-keyword filters, balancing protocol, caption schema, and representative example images---is detailed in Appendix~\ref{app:cocoai}.

\begin{table}[h]
\centering
\small
\caption{Evaluation benchmarks spanning partial-edit hard fakes, 
person-centric AIGC, and multi-person interaction fakes.}
\begin{tabular}{cllll}
\toprule
\# & Dataset & $N$ & Real:Fake & Probes \\
\midrule
1 & Custom Gaze (test) & $4{,}684$ & $2{,}342\!:\!2{,}342$ & partial-edit hard fake \\
2 & COCOAI Person & $15{,}720$ & $2{,}620\!:\!13{,}100$ & person-centric AIGC ($5$ generators) \\
3 & COCOAI Interaction $\bigstar$ & $198$ & $33\!:\!165$ & multi-person interaction (core) \\
\bottomrule
\end{tabular}
\end{table}

Both COCOAI subsets exhibit a $1{:}5$ real-to-fake imbalance under which plain accuracy is uninformative; we therefore report balanced accuracy (BA), macro-F1, and Matthews Correlation Coefficient (MCC), and the unweighted arithmetic mean across the three benchmarks---MCC, which penalizes majority-class collapse, serves as the strictest single indicator of genuine discrimination. All inference runs on a single RTX~4080~SUPER ($16$~GB); FakeVLM variants are loaded in 4-bit NF4 quantization~\citep{dettmers2023qlora} and external baselines at their authors' recommended precision, with greedy decoding ($\textsc{max\_new\_tokens}=64$, SDPA attention) and labels extracted via regex over the first decision keyword.

\paragraph{Baselines.} We compare against seven prior systems spanning low-level forensics and LMM-based detection: FakeVLM origin~\citep{wen2025fakevlm}, Effort~\citep{yan2025effort}, NPR~\citep{tan2024npr}, UnivFD~\citep{ojha2023univfd}, AIDE~\citep{yan2025aide}, SAFE~\citep{li2024safe}, and SIDA-13B~\citep{huang2025sida}, the most direct LMM-based competitor. Per-baseline configurations and the SIDA three-class binarization protocol is detailed in Appendix~\ref{app:baselines}.

\subsection{Main Result: FakeVLM Origin vs.\ Ours}
\label{sec:main-result}

\begin{table}[h]
\centering
\small
\caption{Per-dataset balanced metrics for FakeVLM origin and our BA-best checkpoint, reported as mean $\pm$ std over 3 seeds $\{42,47,53\}$. Best per column in bold.}
\label{tab:main-balanced}
\setlength{\tabcolsep}{3pt}
\resizebox{\linewidth}{!}{%
\begin{tabular}{cccccccccc}
\toprule
\multirow{2}{*}{Method}
& \multicolumn{3}{c}{COCOAI\_Inter}
& \multicolumn{3}{c}{COCOAI\_Person}
& \multicolumn{3}{c}{Custom Gaze} \\
\cmidrule(lr){2-4} \cmidrule(lr){5-7} \cmidrule(lr){8-10}
& BA & F1 & MCC & BA & F1 & MCC & BA & F1 & MCC \\
\midrule
FakeVLM origin     & $68.3{\scriptstyle\pm0.4}$ & $64.9{\scriptstyle\pm0.8}$ & $31.6{\scriptstyle\pm1.2}$
                   & $83.1{\scriptstyle\pm0.1}$ & $78.4{\scriptstyle\pm0.2}$ & $58.4{\scriptstyle\pm0.3}$
                   & $86.5{\scriptstyle\pm0.2}$ & $86.3{\scriptstyle\pm0.3}$ & $75.7{\scriptstyle\pm0.4}$ \\
Ours (mix1650)     & $\mathbf{71.1{\scriptstyle\pm0.3}}$ & $\mathbf{71.9{\scriptstyle\pm0.2}}$ & $\mathbf{43.8{\scriptstyle\pm0.3}}$
                   & $\mathbf{84.3{\scriptstyle\pm0.1}}$ & $\mathbf{83.8{\scriptstyle\pm0.1}}$ & $\mathbf{67.6{\scriptstyle\pm0.2}}$
                   & $\mathbf{99.9{\scriptstyle\pm0.0}}$ & $\mathbf{99.9{\scriptstyle\pm0.0}}$ & $\mathbf{99.8{\scriptstyle\pm0.1}}$ \\
\bottomrule
\end{tabular}}
\end{table}

The pattern admits a clean interpretation. In-distribution (Custom Gaze) saturates at the practical ceiling: BA $86.4 \to 99.9$ ($+13.5$\,pp), macro-F1 $+13.7$\,pp, MCC $+24.4$\,pp. Cross-distribution but person-centric (COCOAI Person, single-person AIGC across five generators including DALL$\cdot$E~3, SDXL, SD3, SD2.1, Midjourney v6) shows consistent moderate gains: BA $+1.3$\,pp, macro-F1 $+5.6$\,pp, MCC $+9.6$\,pp. Cross-distribution and cross-task (COCOAI Interaction)---the diagnostic case, multi-person fully-synthesized AIGC drawn from neither partial edits nor single-person regimes---improves by $+3.7$\,pp BA, $+7.5$\,pp macro-F1, and $+13.3$\,pp MCC; the macro-F1 and MCC margins exceed the BA margin by a factor of two-to-four, the canonical signature of recovering minority-class structure rather than inflating the majority class. The mean improvement is $\mathbf{+6.1}$\,pp BA, $\mathbf{+8.9}$\,pp macro-F1, and $\mathbf{+15.7}$\,pp MCC.

\paragraph{Confusion-matrix dissection on COCOAI Interaction ($n=198$, $33$ real / $165$ fake):}

\begin{table}[h]
\centering
\small
\caption{Confusion-matrix dissection on COCOAI Interaction ($n=198$): 
$+17$ fake recovery with near-zero real-class perturbation.}
\begin{tabular}{lccc}
\toprule
& Origin & Ours & \textbf{$\Delta$} \\
\midrule
\textsc{wrong\_real} (missed fakes / $165$) & $31$ & $14$ & $\mathbf{-17}$ \\
\textsc{right\_fake} (caught fakes / $165$) & $134$ & $151$ & $\mathbf{+17}$ \\
\textsc{right\_real} (real correctly classified / $33$) & $18$ & $17$ & $-1$ \\
\textsc{wrong\_fake} (real misclassified / $33$) & $15$ & $16$ & $+1$ \\
\bottomrule
\end{tabular}
\end{table}

The $+17$-fake recovery is achieved with a near-zero perturbation of real classification ($-1$ \textsc{right\_real}, $+1$ \textsc{wrong\_fake}), refuting the alternative hypothesis that the gain is a calibration drift toward predicting ``fake'': were that the case, real-class accuracy would degrade in proportion to the fake-class gain. The signature---substantial fake recovery without commensurate real degradation---is the hallmark of genuine cue acquisition. The gain profile across the three benchmarks (in-distribution saturation, MCC-driven minority recovery on interaction fakes, modest consistent gains on COCOAI Person) is itself diagnostic of a domain-targeted intervention: the gaze-consistency cue maximally lifts performance on the multi-person interaction regime its theoretical motivation explicitly addresses.

\subsection{External Baseline Comparison}
\label{sec:baselines}

\begin{table}[h]
\centering
\small
\caption{External baseline comparison: balanced metrics across the three benchmarks. Best per column in bold. \emph{Top block}: low-level / classifier baselines and SIDA. \emph{Bottom block}: FakeVLM family (origin vs.\ ours, mix1650).}
\label{tab:baselines-balanced}
\setlength{\tabcolsep}{3pt}
\begin{tabular}{cccccccccc}
\toprule
\multirow{2}{*}{Model}
& \multicolumn{3}{c}{COCOAI\_I}
& \multicolumn{3}{c}{COCOAI\_P}
& \multicolumn{3}{c}{Custom Gaze} \\
\cmidrule(lr){2-4} \cmidrule(lr){5-7} \cmidrule(lr){8-10}
&  BA & macro-F1 & MCC
&  BA & macro-F1 & MCC
&  BA & macro-F1 & MCC \\
\midrule
AIDE              & 48.7 & 28.7 & $-2.2$  & 47.0 & 27.7 & $-5.6$  & 44.0 & 31.5 & $-23.0$ \\
SAFE              & 46.9 & 13.5 & $-22.5$ & 50.0 & 14.3 & 1.1     & 50.0 & 33.3 & 1.4     \\
NPR (nores+tta)   & 51.5 & 17.5 & 7.1     & 50.6 & 17.8 & 2.8     & 28.7 & 22.6 & $-51.4$ \\
UnivFD            & 50.0 & 14.2 & 0.0     & 50.0 & 14.6 & 0.8     & 50.7 & 35.0 & 8.0     \\
SIDA-13B (LMM)    & 60.9 & 57.1 & 17.6    & 68.2 & 66.0 & 32.7    & 83.3 & 82.9 & 69.9    \\
\midrule
FakeVLM origin    & 67.8 & 64.6 & 30.8    & 83.0 & 78.2 & 58.1    & 86.4 & 86.2 & 75.5    \\
\textbf{Ours (mix1650)} & \textbf{71.5} & \textbf{72.1} & \textbf{44.1}
                  & \textbf{84.3} & \textbf{83.8} & \textbf{67.7}
                  & \textbf{99.9} & \textbf{99.9} & \textbf{99.9} \\
\bottomrule
\end{tabular}
\end{table}

The five low-level detectors (AIDE, SAFE, Effort, NPR, UnivFD) collapse to BA $\le 59$ on Interaction and $\le 71$ on Person, with NPR's and UnivFD's MCC at-or-near zero across all three benchmarks---the low-level paradigm scarcely beats chance against contemporary multi-generator person-centric content (per-model breakdown and qualitative limitation analysis in Appendix~\ref{app:baselines}). SIDA-13B, the most direct LMM competitor at $13$B parameters, attains a respectable mean BA of $70.8$ but is surpassed by our $7$B model by $\mathbf{+14.4}$\,pp on unweighted mean BA, $+16.6$\,pp on macro-F1, and $+30.5$\,pp on MCC. Two findings follow. Within the LMM-detector family, supervision design dominates parameter count: a $7$B model with deliberately constructed gaze supervision exceeds a $13$B model trained on conventional fake-detection corpora across every metric on every benchmark. The $+6.1$\,pp jump from FakeVLM origin (mean BA $79.0$) to ours (mean BA $85.2$) is achieved under identical backbone and inference protocol, isolating data-and-caption design as the locus of the gain.

The advantage widens under MCC. Ours achieves mean MCC $70.5$, the only model in the comparison to clear the $70$ mark; SIDA reaches $40.0$ ($\Delta$ $-30.5$\,pp), FakeVLM origin $54.8$ ($\Delta$ $-15.7$\,pp), and the five low-level detectors all fall below $18$, three of them below zero. Among the eight systems compared, only ours produces predictions whose joint confusion structure clears the majority-class baseline by a non-trivial margin on every benchmark.

\subsection{Cross-Architecture Generalization}
\label{sec:cross-arch}

We test whether the gaze cue is FakeVLM-specific or transferable to other visual backbones. We fine-tune Effort~\citep{yan2025effort}---a vision-only CLIP-ViT-L/14 detector with SVD-residual self-attention that shares only the CLIP backbone with FakeVLM---on the $37{,}464$-sample Custom Gaze training split under a DeepfakeBench~\citep{yan2023deepfakebench}-style recipe. Because Effort emits a scalar score rather than a caption, supervision necessarily reduces to label level; what transfers is therefore the paired-edit data structure and the gaze cue itself, not the caption schema. Table~\ref{tab:effort-family} reports this canonical variant (Ours, Effort gaze-FT) alongside Effort origin, while a parallel UFD-style recipe exhibits the same in-distribution saturation and same-direction cross-distribution transfer (recipe details and per-benchmark numbers in 
Appendix~\ref{app:hparams-effort}).

\begin{table}[h]
\centering
\small
\caption{Effort family: cross-architecture validation on a vision-only Effort backbone (CLIP-ViT-L/14 with SVD-residual self-attention; shares only the CLIP backbone with FakeVLM). \emph{Origin}: Effort released checkpoint; \emph{Ours, Effort gaze-FT}: Effort fine-tuned on the $37{,}464$-sample Custom Gaze training split under a DeepfakeBench-style recipe (label-level supervision; \S\ref{sec:cross-arch}, Appendix~\ref{app:hparams-effort})}
\label{tab:effort-family}
\setlength{\tabcolsep}{3pt}
\begin{tabular}{cccccccccc}
\toprule
\multirow{2}{*}{Model}
& \multicolumn{3}{c}{COCOAI\_I}
& \multicolumn{3}{c}{COCOAI\_P}
& \multicolumn{3}{c}{Custom Gaze} \\
\cmidrule(lr){2-4} \cmidrule(lr){5-7} \cmidrule(lr){8-10}
&  BA & macro-F1 & MCC
&  BA & macro-F1 & MCC
&  BA & macro-F1 & MCC \\
\midrule
Effort origin                    & 59.0 & 49.9 & 13.5 & 70.6 & 58.2 & 30.8 & 50.9 & 36.1 & 6.7  \\
\textbf{Ours (Effort, gaze-FT)}  & \textbf{62.4} & \textbf{58.1} & \textbf{20.0}
                                 & \textbf{75.2} & \textbf{66.2} & \textbf{39.5}
                                 & \textbf{99.9} & \textbf{99.9} & \textbf{99.9} \\
\bottomrule
\end{tabular}
\end{table}

Effort gaze-FT saturates Custom Gaze at the same ceiling as FakeVLM ($+49.0$\,pp BA, $+93.2$\,pp MCC; Table~\ref{tab:baselines-balanced}) despite a fundamentally different architecture, and on the COCOAI subsets reproduces FakeVLM's minority-class-recovery signature with macro-F1 and MCC gains exceeding BA gains on every benchmark.

We synthesize the empirical evidence into a four-step mechanistic account:

\textbf{(M1) Paired-edit shortcut blocking.} All $18{,}732$ training pairs (Custom Gaze train split; $23{,}415$ across the full dataset) share base identity, scene, and lighting; the only intra-pair difference is the eye region. The detector cannot reduce loss by latching onto FLUX-specific global fingerprints; the optimization landscape forces anatomically grounded eye-region features.

\textbf{(M2) CLIP prior preservation.} Both Effort (SVD-residual) and our LoRA configuration preserve the bulk of CLIP's pretrained representations~\citep{radford2021clip}. Cross-generator generalization is inherited from CLIP's $400$M-pair pretraining---FLUX, SDXL, SD3, and DALL$\cdot$E imagery occupy approximately commensurable feature regions---rather than learned by the detector in any narrow sense.

The decisive empirical signature is the cross-architecture pattern itself: a non-LMM vision-only detector, trained on identical supervision, exhibits the same in-distribution saturation and the same direction of cross-distribution transfer. Were the cue an architecture-specific phenomenon, this replication would not occur.

\section{Ablation Study}
\label{sec:ablation}

Holding backbone, LoRA configuration, training mixture, optimizer, and checkpoint-selection protocol constant, we compare two endpoints of caption-supervision depth: the proposed five-sentence Block-Compositional schema (full reasoning skeleton, $1{,}250$ unique captions) versus a one-sentence decision-only schema (minimal label-statement, $2$ unique captions). This A-vs-B contrast isolates the marginal contribution of the reasoning skeleton.

\begin{table}[h]
\centering
\small
\caption{Supervision depth variants: 5-block schema (1,250 captions) 
vs.\ decision-only baseline (2 captions) isolating the reasoning skeleton.}
\resizebox{0.83\linewidth}{!}{%
\begin{tabular}{lccc}
\toprule
Variant & Caption sentences & Positions used & Unique captions \\
\midrule
A. Mix (proposed) & $5$ & $p_1 + p_2 + p_3 + p_4 + p_5$ & $1{,}250$ \\
B. Decision-only & $1$ & $p_1$ & $2$ \\
\bottomrule
\end{tabular}%
}
\end{table}

\begin{table}[h]
\centering
\small
\caption{Caption design ablation. Each variant evaluated at its own BA-best checkpoint, selected on the same 970-sample dev mixture as the deployed model. The COCOAI Inter / Person columns report fake-class accuracy on the deepfake-only subsets ($n{=}270$ for Interaction and $n{=}1{,}000$ for Person), chosen so that the caption-design contrast is read directly off fake-class detection performance without dilution from the real class; the Custom Gaze column reports balanced accuracy on the $4{,}684$-sample 1:1 split. Mean is the unweighted arithmetic mean across the three displayed benchmarks.}
\setlength{\tabcolsep}{5pt}
\resizebox{0.83\linewidth}{!}{%
\label{tab:ablation-caption}
\begin{tabular}{lcccc}
\toprule
Variant & COCOAI\_I & COCOAI\_P & Custom Gaze & Mean \\
\midrule
A. Mix (5-sent, st1650) & $\mathbf{84.3}$ & $88.1$ & $99.9$ & $\mathbf{90.7}$ \\
B. Decision-only (1-sent, st1500) & $81.4$ & $\mathbf{90.7}$ & $99.9$ & $90.6$ \\
\bottomrule
\end{tabular}%
}
\end{table}

At first glance the result is paradoxical: the means differ by only $+0.1$\,pp ($A=90.7$ vs.\ $B=90.6$), yet the per-benchmark distribution is non-trivial. A leads by $\mathbf{+2.9}$\,pp on COCOAI Interaction (our core-evidence benchmark), B leads by $+2.6$\,pp on COCOAI Person, and both saturate on Custom Gaze. The opposite-sign deltas cancel in the mean.

\paragraph{Interpretation.} Reasoning depth is \emph{not} a uniform enhancer; it is a domain-targeted lever whose effect manifests only where the reasoning aligns with the benchmark's semantic structure. On Custom Gaze, the base classification signal saturates without any reasoning. On COCOAI Person, a five-sentence rationale built around multi-person gaze geometry over-specializes for single-person fakes. Only on multi-person interaction fakes---where the gaze skeleton aligns by design---does additional reasoning depth produce a positive gain. The opposite-sign deltas are consistent with a domain-targeted interpretation, and are invisible in the aggregate Mean. Card-level failure analysis, mode-collapse diagnostics, and the inference-time behavior of variant B (confirming that reasoning capacity is acquired through supervision) appear in Appendices~\ref{app:cards}, \ref{app:three-variant}, and \ref{app:negative}.

\section{Discussion and Limitations}
\label{sec:discussion}

Our results admit a two-axis cue decomposition reading: gaze geometry, head--eye alignment, and attentional plausibility form a high-level semantic-consistency axis complementary to the low-level paradigms (NPR, UnivFD, AIDE, Effort origin) whose territory contemporary generators progressively reclaim. The scope is deliberately person-centric; we delineate four limitations---(L1) domain specialization, (L2) card over-trust, (L3) single-generator (FLUX.1-Fill) construction, and (L4) face-detection dependency at training time---and three future directions (video extension, multi-cue composition, principled card gating) in Appendix~\ref{app:discussion}.

\section{Conclusion}

We have argued that AI-generated image detection, in the current generator regime, requires an additional cue axis beyond the low-level paradigm---one grounded in social-semantic consistency rather than pixel statistics. Operationalizing this argument through Social Gaze Consistency, Block-Compositional Caption Supervision, and the Custom Gaze paired-edit dataset, we have demonstrated, on three person-centric benchmarks: (i) on the COCOAI benchmark, balanced-accuracy gains of $\mathbf{+3.7}$\,pp on the interaction-centric subset ($67.8 \to 71.5$) and $\mathbf{+1.3}$\,pp on the person-centric subset ($83.0 \to 84.3$) over FakeVLM origin, with corresponding leads of $\mathbf{+10.6}$\,pp and $\mathbf{+16.1}$\,pp BA over a $13$B LMM competitor (SIDA-13B), achieved with a $7$B backbone via supervision design alone; (ii) cross-architecture transfer to a vision-only Effort backbone, with parallel in-distribution saturation ($+49.0$\,pp BA on Custom Gaze) and consistent same-direction cross-distribution gains on every metric, confirming the cue's architecture-agnostic nature; (iii) a domain-targeted reasoning-depth effect isolable only by examining benchmark-specific rather than averaged behavior---a $+2.9$\,pp lift on multi-person interaction fakes that fully cancels in the mean against the opposite-sign delta on single-person fakes; and (iv) a four-step mechanistic account---paired-edit shortcut blocking, hard-to-easy difficulty transfer, CLIP prior preservation, and diffusion-family shared spectral weakness in periocular structure---explaining why single-generator inpainting supervision transfers to multi-generator full-image evaluation across architectures. We expect the semantic-consistency cue space to become increasingly central as low-level artifacts continue to recede.



{
\small
\bibliography{reference}



}


\appendix
\newpage

\newpage
\appendix

\section{Datasets: Custom Gaze and COCOAI}
\label{app:datasheet}

This appendix provides datasheets for both datasets used in the paper:
the proposed \emph{Custom Gaze} and the externally-curated
\emph{COCOAI} suite (Person and Interaction subsets).

\subsection{Custom Gaze Datasheet}
\label{app:custom-gaze}

We follow the Datasheets-for-Datasets framework~\citep{gebru2021datasheets}.

\paragraph{Datasheet summary.}
\begin{table}[h]
\centering\small
\begin{tabular}{ll}
\toprule
\textbf{Field} & \textbf{Value} \\
\midrule
Name                          & Custom Gaze \\
Total samples                 & $46{,}830$ ($23{,}415$ real / $23{,}415$ fake; $1\!:\!1$ paired)\\
Real source                   & Open Images Mutual Gaze (OI-MG)~\citep{doosti2021oimg}\\
Fake construction             & FLUX.1-Fill-dev~\citep{bfl2024flux} eye-region inpainting\\
Mask region                   & $[y_{\min}\!+\!0.25h,\,y_{\min}\!+\!0.55h]$ within face bbox\\
Pair structure                & Identity-preserved (real/fake share base ID)\\
Splits (train\,:\,val\,:\,test) & $8\!:\!1\!:\!1$ ($37{,}464\!:\!4{,}682\!:\!4{,}684$)\\
Caption schema                & Block-Compositional, $|\mathcal{C}|=1{,}250$ unique\\
Image license                 & CC-BY-NC-4.0\\
Caption-pool license          & CC0\\
\bottomrule
\end{tabular}
\end{table}

\paragraph{Real source filtering.}
OI-MG contains binary mutual-gaze annotations on $29{,}228$ Open Images.
We retain only $\textsc{Annotation}=1$ samples (mutual eye contact):
$7{,}126$ retained from $26{,}410$ train candidates ($\approx\!27.0\%$),
$1{,}846$ retained from $6{,}659$ test candidates ($\approx\!27.7\%$).
Multiple face-bounding-box pairs per image generate distinct pair candidates,
yielding $23{,}415$ effective real pairs. Each retained sample carries two
face bounding boxes
$(x_{\min},y_{\min},x_{\max},y_{\max})_{A,B}$;
the \emph{fake\_A} variant perturbs participant A's eye region only.

\paragraph{Upstream licensing of OI-MG.}
The OI-MG mutual-gaze \emph{annotations} follow the
upstream Doosti et al.~\citep{doosti2021oimg} terms; 
the underlying photographs are the
Open Images V6 subset, on which Open Images itself attaches a
CC-BY-2.0 licence to the dataset metadata while individual photographs
retain their original Flickr per-image licences (typically CC-BY-2.0,
CC-BY-SA-2.0, or CC-BY-NC-2.0, recorded per-image in the Open Images
distribution). The Custom Gaze derivative carries a
CC-BY-NC-4.0 licence (more restrictive than, and compatible with, the most
permissive Open Images per-image licence), with the Open Images
per-image licence pointers retained in the per-image metadata JSONs
so that downstream users can audit per-image upstream terms.

\paragraph{Eye-region mask.}
Given a face bounding box,
\begin{equation}
M_{\mathrm{eye}} = \left\{ (i, j) \;\middle|\;
\begin{aligned}
& y_{\min} + 0.25\,h \;\leq\; i \;\leq\; y_{\min} + 0.55\,h, \\
& x_{\min} + 0.05\,w \;\leq\; j \;\leq\; x_{\max} - 0.05\,w
\end{aligned}
\right\},
\end{equation}
with $h=y_{\max}-y_{\min}$, $w=x_{\max}-x_{\min}$. Mask boundaries are
softened by Gaussian blur of radius~$3$\,px on the alpha channel.
The vertical band $[0.25h,0.55h]$ was selected by a $200$-sample manual
sweep over four candidate vertical bands $\{[0.20,0.50],[0.25,0.55],[0.30,0.60],[0.20,0.55]\}$, where
$[0.25,0.55]$ uniquely preserved skin-tone continuity at the upper boundary
while fully covering iris and eyelid in $>95\%$ of the $200$ inspected cases.

\paragraph{FLUX.1-Fill inpainting pipeline.}
\begin{verbatim}
from diffusers import FluxFillPipeline
import torch
pipe = FluxFillPipeline.from_pretrained(
    "black-forest-labs/FLUX.1-Fill-dev", torch_dtype=torch.bfloat16)
pipe.enable_model_cpu_offload(); pipe.enable_attention_slicing()
prompt = ("eyes looking to the side, avoiding eye contact, "
          "natural eye movement, same person, photorealistic")
num_inference_steps = 20      # deployed value (qualitatively similar over 20--28)
guidance_scale       = 25     # deployed value (qualitatively similar over 25--30)
# Image preprocessing aligns to the FLUX 16-px grid:
max_size = 1024
ratio    = max_size / max(w, h)
new_w    = (int(w * ratio) // 16) * 16
new_h    = (int(h * ratio) // 16) * 16
\end{verbatim}
The prompt is compositional: \emph{``eyes looking to the side, avoiding
eye contact''} specifies the gaze perturbation, \emph{``natural eye
movement''} biases away from anatomical extremes, and
\emph{``same person, photorealistic''} preserves identity. We deliberately
avoid lexical tokens such as \emph{``fake''}, \emph{``edit''}, or
\emph{``mask''} because pilot runs showed these tokens degraded local
photorealism (visibly blurred iris boundaries and reduced specular-highlight
coherence on the cornea). Per-image generation takes
$7$--$10$\,s at $1024$-px max side and $20$ inference steps on RTX A6000;
the BF16+CPU-offload+attention-slicing configuration fits within
$16$\,GB VRAM, allowing single-GPU dataset construction.

\paragraph{Block-Compositional caption schema (full macro pool).}
Each Custom Gaze sample carries a $5$-tuple caption
$c=(p_1,p_2,p_3,p_4,p_5)$ with positions
\textsc{Decision}/\textsc{Scene}/\textsc{Method}/\textsc{Evidence}/\textsc{Conclusion}.
$p_1$ is label-deterministic (\emph{``This is a (real$\mid$fake) image.''}).
$p_2,p_3$ are label-shared, each with $5$ surface variants;
$p_4,p_5$ are label-branched with $5$ variants per label.
The cardinality is
\begin{equation}
|\mathcal{C}| \;=\; \underbrace{1}_{p_1} \cdot \underbrace{5}_{p_2}
                 \cdot \underbrace{5}_{p_3} \cdot \underbrace{5}_{p_4}
                 \cdot \underbrace{5}_{p_5} \cdot \underbrace{2}_{\text{label}}
\;=\; 1{,}250\quad\text{unique captions.}
\end{equation}
Representative variants:
\begin{itemize}[leftmargin=2em]
\item $p_2$: \emph{``As more than one person appears in the image, I focused
primarily on gaze behavior and eye-region consistency.''} (5 paraphrases.)
\item $p_3$: \emph{``I evaluated the image mainly by checking whether the
gaze positions match plausible interaction targets and whether the pupils
appear naturally aligned across both eyes.''} (5 paraphrases.)
\item $p_4$ (real): \emph{``The eye contact and gaze targets look
believable, with left-right eye alignment remaining visually consistent.''}
\item $p_4$ (fake): \emph{``The eye contact looks implausible for the
scene, and the left-right eye alignment appears unstable.''}
\item $p_5$ (real): \emph{``Overall, the gaze and eye alignment appear
natural enough to support a real image.''}
\item $p_5$ (fake): \emph{``Overall, the gaze and eye alignment appear
unnatural enough to suggest a fake image.''}
\end{itemize}
Every one of the $1{,}250$ captions follows the same Decision $\to$ Scene
$\to$ Method $\to$ Evidence $\to$ Conclusion ordering; only surface
lexicalisation varies. This is the operationalisation of
\emph{single reasoning path $\times$ multi-surface}.

\begin{figure}[!htb]
\centering
\includegraphics[width=\linewidth]{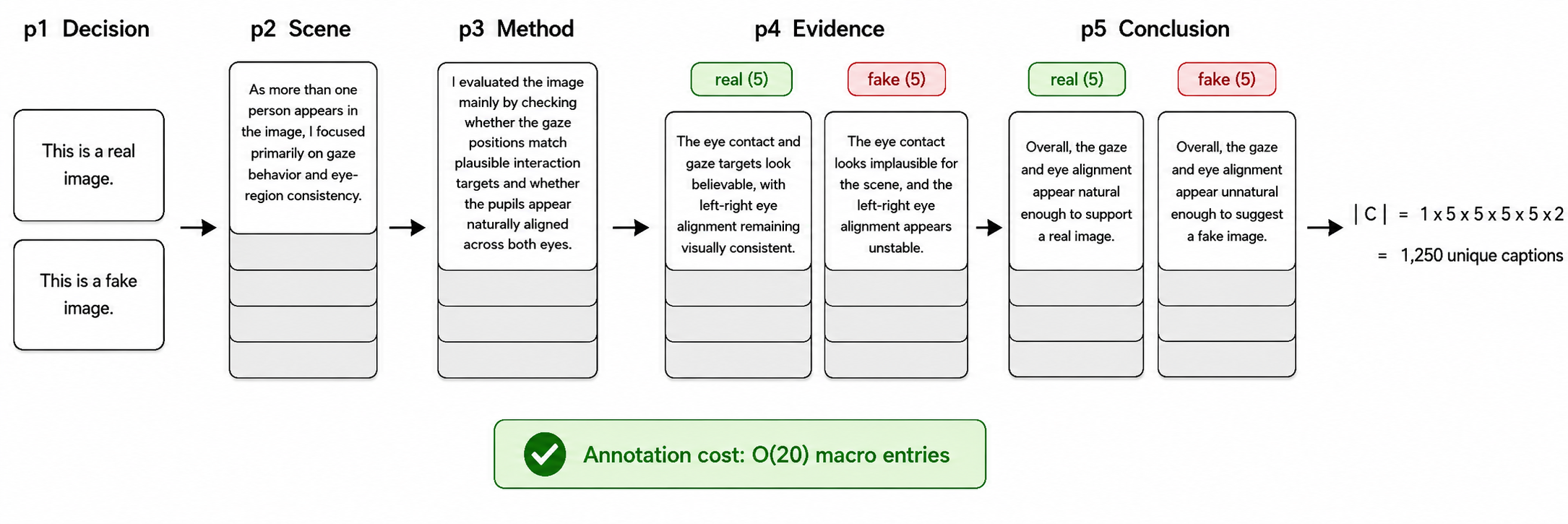}
\caption{Block-Compositional Caption schema.}
\label{fig:caption_schema_compositional}
\end{figure}

\paragraph{Splits and pair-level grouping.}
Splits are formed by treating each base ID as an atomic unit, then
unpacking into individual images. Stratification preserves
$1\!:\!1$ real-fake balance per split; random seed $42$.
\begin{table}[h]
\centering\small
\caption{Custom Gaze splits. Pair count equals real count equals fake count by construction (each pair contributes exactly one real and one fake image); Total $=$ Real $+$ Fake $=$ $2 \times$ Pairs.}
\label{tab:custom-gaze-splits}
\begin{tabular}{lcccc}
\toprule
Split & Pairs & Real & Fake & Total \\
\midrule
Train ($80\%$) & $18{,}732$ & $18{,}732$ & $18{,}732$ & $37{,}464$ \\
Val ($10\%$)   & $2{,}341$  & $2{,}341$  & $2{,}341$  & $4{,}682$ \\
Test ($10\%$)  & $2{,}342$  & $2{,}342$  & $2{,}342$  & $4{,}684$ \\
\midrule
Total          & $23{,}415$ & $23{,}415$ & $23{,}415$ & $46{,}830$ \\
\bottomrule
\end{tabular}
\end{table}
Pair-level grouping is essential for the shortcut-blocking mechanism: with
identity-grouped splits, every val/test base ID has zero training-time
exposure under either label, so an identity-memorisation shortcut cannot
emerge.

\paragraph{Per-model effective sample counts.}
LMMs occasionally emit outputs that the regex parser
(\texttt{\textasciicircum This is a (real$\mid$fake) image\textbackslash.})
cannot match; such samples are excluded rather than coerced.
\begin{table}[h]
\centering\small
\begin{tabular}{lccc}
\toprule
Model & Custom Gaze & COCOAI Person & COCOAI Inter \\
\midrule
FakeVLM origin           & $4{,}681$ ($-3$)  & $15{,}720$ ($0$) & $198$ ($0$) \\
Ours mix1650 (FakeVLM)   & $4{,}676$ ($-8$)  & $15{,}713$ ($-7$) & $198$ ($0$) \\
SIDA-13B                 & $4{,}684$ ($0$)  & $15{,}720$ ($0$) & $198$ ($0$) \\
Effort gaze-FT (BR-2ep)  & $4{,}684$ ($0$)  & $15{,}720$ ($0$) & $198$ ($0$) \\
\bottomrule
\end{tabular}
\end{table}
Parsing-failure rates remain $<0.06\%$ on every benchmark. Vision-only
detectors emit scalar scores and have $n_{\text{eff}}\!=\!n_0$ by construction.

\paragraph{Licensing.}
The Custom Gaze image set is licensed under CC-BY-NC-4.0 (compatible
with the OI-MG upstream CC-BY licence inherited via Open Images), with
commercial use disallowed without explicit author consultation; the
caption macro pool is licensed under CC0. The licence terms restrict
use of both the dataset and detector checkpoints to non-commercial use
under CC-BY-NC-4.0 and explicitly forbid any deployment that
misrepresents a depicted individual's actual gaze, attention, or
interaction.

\paragraph{Dual-use and broader impacts.}
The detector and dataset are dual-use. \emph{Positive impacts.} A robust
person-centric AIGC detector supports identity protection of depicted
individuals against unauthorised partial-edit manipulation, and
disinformation defence in person- and interaction-centric domains where
existing low-level detectors (NPR, UnivFD, AIDE, SAFE, Effort origin)
collapse on contemporary multi-generator content (cf.\ Table~\ref{tab:baselines-balanced}).
\emph{Negative impacts and mitigations.} (i) Mis-deployment that
misrepresents a depicted individual's actual gaze, attention, or
interaction is forbidden by the licence clause above. (ii)
Cross-\emph{inpainter} transfer (SDXL-Inpaint, IP-Adapter, MGIE, etc.)
is not directly verified (limitation~L3); adversaries adopting non-FLUX
inpainters may evade detection until cross-inpainter transfer is
empirically established. (iii) The deployed checkpoint exhibits
single-person card over-trust on COCOAI Person ($\S$\ref{app:negative} NR2)
and is therefore unsuitable for high-stakes single-person verification
without additional gating; the principled card-gating remedy outlined in
$\S$\ref{app:disc-limits} is recommended before any production use.

\paragraph{Consent provenance.}
Custom Gaze is built on the publicly available OI-MG corpus, whose
photographs are the Open Images V6 subset distributed under per-image
Flickr licences (CC-BY-2.0 / CC-BY-SA-2.0 / CC-BY-NC-2.0 family). We
did \emph{not} seek additional consent from depicted individuals beyond
this upstream licensing flow, because (a) the photographs are already
publicly distributed under licences that explicitly permit redistribution
and modification (within CC-BY family), (b) the Custom Gaze derivative
carries a more restrictive CC-BY-NC-4.0 downstream licence plus a
deployment-misrepresentation prohibition, and (c) the Open Images
per-image licence pointers are retained in the per-image metadata JSONs
so that downstream users can audit and respect each subject's upstream
terms.

\paragraph{Limitations and ethical considerations.}
(i) Identifiable-individual imagery; redistribution complies with
Open Images licensing, and the license clause forbids any deployment
that misrepresents a depicted individual's actual gaze, attention, or
interaction. (ii) Single-generator (FLUX.1-Fill) construction;
cross-generator transfer is empirically demonstrated in the main
paper's \S\ref{sec:cross-arch}; cross-inpainter transfer is not directly
verified.
(iii) Mutual-gaze-positivity restriction: non-mutual-gaze imagery is
out of scope. (iv) OI-MG inherits the geographic and demographic biases inherent to Open Images; downstream deployments should audit population coverage.

\subsection{COCOAI Person and COCOAI Interaction Construction}
\label{app:cocoai}

The COCOAI suite~\citep{roy2026cocoai} is a person-centric AIGC
evaluation set assembled from COCO \texttt{val2017} source imagery;
we report results on two of its subsets. Both subsets share an identical first-stage
\emph{caption-keyword filter}---a deliberate design choice that selects
images whose COCO caption contains at least one of the eight
unambiguously human nouns
\{\texttt{man}, \texttt{men}, \texttt{woman}, \texttt{women},
\texttt{people}, \texttt{person}, \texttt{boy}, \texttt{girl}\}.
We deliberately exclude collective nouns such as \emph{group},
\emph{family}, or \emph{crowd}: COCO captions use these terms to
describe non-human assemblies (e.g., ``a group of zebras'') with
non-trivial frequency, and including them admits animal-only or mixed
animal--human imagery that would contaminate a person-centric
benchmark. The eight retained nouns are \emph{strictly} human.

\paragraph{COCOAI Person (single-person fake suite).}
After the caption-keyword filter, candidate images are paired with
their five generator-rendered counterparts (DALL$\cdot$E~3, SDXL,
SD3, SD2.1, Midjourney~v6) released by Roy et al.~\citep{roy2026cocoai}, and
real/fake assignment is recorded under the dataset's
\texttt{person\_1200} schema. The paired set used in this
paper contains $n=15{,}720$ samples ($2{,}620$ real $\cup\,
5\!\times\!2{,}620$ generator-paired fakes); the larger
$2{,}620$-real pool is drawn from a $\approx\!2{,}600$-image
caption-filtered candidate set, with additional reals available for
extension if needed. Per-generator slices are exactly $2{,}620$
fakes paired with the $2{,}620$-real shared half.

\paragraph{COCOAI Interaction (multi-person interaction subset).}
The interaction subset is constructed by a four-step procedure on
top of the caption-keyword filter:
(1) the same eight-noun caption filter as COCOAI Person selects
candidate images;
(2) a second-pass de-duplication removes any candidate whose caption
overlaps with captions already extracted in an earlier filtering
round (preventing caption-level leakage across subsets);
(3) the surviving candidates are partitioned by the dataset's
\texttt{label\_b} field, which records the generator that produced each
fake (real samples form the dedicated \texttt{label\_b}=real partition);
(4) within each \texttt{label\_b} partition, the count is balanced down
to the smallest partition's size ($\texttt{label\_b\_1}=2{,}620$
in the dataset's native counting), yielding the curated interaction
benchmark. The interaction view used throughout this paper
contains $n=198$ samples ($33$ real, $165$ fake), the
$\texttt{label\_b}$-balanced subset reserved for evaluation. The
skew toward fake is intentional: the benchmark asks \emph{can the
detector recognize multi-person fake content without relying on
real-class statistics}, which is the regime in which gaze geometry
is forensically informative.

\paragraph{Two-view evaluation protocol.}
The COCOAI test JSONs distributed with each baseline use a fake-only split
($n=270$ for Interaction, $n=1{,}000$ for Person) where real-class accuracy
is undefined and balanced accuracy degenerates to fake-class accuracy.
The paired view (used in this paper's main result table) takes the full
$n=198$ / $n=15{,}720$ sets. The paired and fake-only views are
\emph{not interchangeable}; cross-walk constants for FakeVLM origin are
$\text{BA}_{\text{paired}}=67.8$ vs $f\_acc_{\text{fake-only}}=75.6$ on
COCOAI Inter, and $83.0$ vs $84.9$ on COCOAI Person (for ours mix1650:
$71.5$ vs $84.4$ and $84.3$ vs $88.1$). The paired view is canonical
throughout this paper; the fake-only $f\_acc$ view appears in the
caption-design ablation tables in order to read the A vs B contrast
directly off fake-class accuracy, where the supervision-design effect
on detecting fakes is most cleanly observed.

\paragraph{Caption / supervision schema on COCOAI.}
COCOAI ships only binary fake/real labels on the paired sets; no
caption-style annotation is released. The detector's caption output on
COCOAI is therefore the model's own learned $5$-block reasoning
under the Custom Gaze supervision, applied at inference time to images
the model never trained on. This makes COCOAI a clean cross-distribution
test of whether the supervision-injected reasoning skeleton transfers to
imagery whose pixel statistics (camera, post-processing, generator pipeline)
differ from the FLUX.1-Fill training distribution.

\paragraph{Per-generator size accounting.}
Every per-generator slice on COCOAI Person is $5{,}240$ samples
(the $2{,}620$-real half is shared across the $5$ generators, paired
with each generator's own $2{,}620$ fakes); per-generator balanced
accuracy is computed on these $5{,}240$-sample slices and is consumed by
the cross-architecture analysis in \S\ref{sec:cross-arch}.

\section{Training Setup, Per-Baseline Configurations, and Inference Protocol}
\label{app:hparams}
\label{app:baselines}

This section consolidates training and evaluation configurations.
We report the FakeVLM mixed fine-tune, the two Effort recipes,
the per-baseline inference protocol, and a qualitative limitation
analysis of every external baseline.
Values marked $\bullet$ are verified against \texttt{trainer\_state.json}
or per-recipe eval logs; $\circ$ are present in the launch script and
re-extracted from the original training command.

\subsection{FakeVLM Mixed Fine-Tuning Recipe}
\label{app:hparams-fakevlm}

The deployed checkpoint is \texttt{lingcco/fakeVLM} (LLaVA-v1.5-7B fully
tuned on FakeClue) further fine-tuned on a $151{,}173$-sample mixture
($46{,}830$ Custom Gaze $+$ $104{,}343$ FakeClue, FakeClue half retaining
its original GPT free-form captions). LoRA adapters are attached to all
seven LLM projection matrices (\texttt{q,k,v,o,gate,up,down\_proj}) and to
every \texttt{Linear} reachable under \texttt{vision\_tower}; the 2-layer
MLP projector is full-FT via \texttt{modules\_to\_save}.

\begin{table}[h]
\centering\caption{FakeVLM mixed fine-tuning recipe.}\label{tab:hparam-fakevlm}\small
\resizebox{\linewidth}{!}{
\begin{tabular}{@{}lll@{}}
\toprule
Field & Value & Source \\
\midrule
Base / vision / LLM             & lingcco/fakeVLM ; CLIP-ViT-L/14-336 ; Vicuna-v1.5-7B & $\bullet$ \\
LoRA $r$, $\alpha$, dropout, bias & $r{=}16$, $\alpha{=}32$, $0.05$, none           & $\circ$ \\
Trainable params                 & $\approx 45$M of $\approx 7$B ($\approx 0.6\%$) & derived \\
Optimizer                        & AdamW, $\beta\!=\!(0.9,0.999)$, $\lambda\!=\!0$  & $\circ$ HF default \\
Peak LR / schedule               & $2{\times}10^{-5}$ ; cosine + warmup ratio $0.03$ & $\bullet$ \\
Per-device batch / grad-accum / world & $1$ / $4$ / $8$ (RTX A6000 48GB)            & $\circ$ \\
Effective batch                  & $32$                                              & derived \\
Precision / DeepSpeed            & BF16 + TF32 ; ZeRO-2 & $\circ$ \\
Flash-attn / grad-ckpt           & off / off                                         & $\circ$ \\
Seq.\ max length                 & $1{,}024$ tokens                                  & $\circ$ \\
Mask question tokens             & True (loss on assistant turn only)                & $\circ$ \\
Epochs / steps                   & $2$ ; \texttt{max\_steps}$=7{,}558$, end at $7{,}550$ & $\bullet$ \\
Eval / save / log interval       & $50$ / $50$ / $20$ steps                          & $\bullet$ \\
RNG seed                         & $42$                                              & $\circ$ \\
In-training dev set              & $970$ ($470$ Gaze $+$ $500$ FakeClue test mix)    & $\bullet$ \\
\texttt{metric\_for\_best\_model} & \texttt{eval\_balanced\_accuracy}, greater-is-better & $\circ$ \\
Best metric / step               & $0.9990$ at step $1{,}650$                        & $\bullet$ \\
Loss-min step (decoupling)       & step $2{,}850$, $\mathcal{L}=0.2252$              & $\bullet$ \\
Total FLOPs                      & $1.039{\times}10^{19}$                            & $\bullet$ \\
\bottomrule
\end{tabular}}
\end{table}

\paragraph{BA / loss decoupling.}
The $1{,}200$-step gap between BA-best (step $1{,}650$, $\mathrm{BA}=0.9990$)
and loss-min (step $2{,}850$, $\mathcal{L}=0.2252$) is consequential.
Under token-level cross entropy with template-uniformity-tolerant decoding,
$\arg\min_\theta \mathcal{L}\neq\arg\max_\theta \mathrm{BA}$, and selecting
on $\mathcal{L}$ alone forfeits $\sim 0.0021$ in BA on the $970$-sample
dev mix. The deployed checkpoint is $\theta_{1650}$.

\paragraph{Checkpoint composition.}
The released FakeVLM checkpoint contains only the LoRA adapter and
projector full-FT deltas ($\sim\!45$M trainable parameters); base
weights are obtained from the upstream HuggingFace repository
(\texttt{lingcco/fakeVLM}) under the original LLaMA-2 / LLaVA / CLIP
licences. The two Effort variants follow the same convention, releasing
only the trained residual-SVD blocks and the linear head.

\subsection{Effort Fine-Tuning Recipes}
\label{app:hparams-effort}

We fine-tune Effort~\citep{yan2025effort} under two recipes---a UFD-style
recipe and a DeepfakeBench~\citep{yan2023deepfakebench}-style recipe---that
share the same backbone but differ in optimization regime. Both freeze the
top-$1{,}023$ singular vectors of every attention matrix as \texttt{weight\_main}
and train only the residual $U/S/V$ blocks plus a from-scratch
$\mathrm{Linear}(1024\!\to\!1)$ head.

\begin{table}[h]
\centering\caption{Effort fine-tuning recipes.}\label{tab:hparam-effort}\small
\begin{tabular}{@{}lll@{}}
\toprule
Field & UFD recipe & DeepfakeBench recipe \\
\midrule
Backbone / weights         & CLIP-ViT-L/14 / openai/clip-vit-large-patch14 & idem \\
SVD residual rank          & $1{,}023$ frozen $+$ $1$ trained               & idem \\
Loss / optimizer           & BCE-with-logits / AdamW, $\beta\!=\!(0.9,0.999)$, $\lambda\!=\!0$ & idem \\
LR                         & $2{\times}10^{-4}$ constant (no schedule)      & $2{\times}10^{-4}$, recipe default \\
Train data                 & Custom Gaze train $37{,}464$                   & idem \\
\texttt{loadSize / cropSize} & $256$ / $224$                                & idem \\
Train / eval transform     & RandomCrop$(224)$+HFlip / CenterCrop$(224)$    & idem \\
Augmentation (blur/jpeg)    & off / off                                      & idem \\
Batch / niter / patience / seed & $32$ / $2$ / $5$ / $0$                    & idem \\
Hardware                   & $1\times$ RTX 4080 SUPER 16GB                  & idem \\
Selected checkpoint (best val.\ BA) & epoch $1$ (of $2$)                   & epoch $0$ (of $2$) \\
\bottomrule
\end{tabular}
\end{table}

\subsection{Per-Baseline Inference Protocol and Limitation Analysis}
\label{app:baselines-detail}

External baselines run at their authors' recommended precision and threshold.
LMM baselines (FakeVLM origin, ours, SIDA-13B) decode greedily with
\texttt{max\_new\_tokens}$=64$ and SDPA attention; the FakeVLM family is
loaded in 4-bit NF4 quantization to fit within the inference GPU's
$16$\,GB envelope (applied identically to FakeVLM origin and ours so that
the comparison isolates supervision design from quantization effects).

\begin{table}[h]
\centering\caption{Per-baseline inference protocol.}\label{tab:hparam-inference}\footnotesize
\begin{tabular}{@{}p{0.20\textwidth}p{0.34\textwidth}p{0.10\textwidth}p{0.22\textwidth}@{}}
\toprule
Baseline & Backbone / weights & Precision & Decoding / threshold \\
\midrule
AIDE~\citep{yan2025aide}              & ResNet-50 / \texttt{GenImage\_train.pth}                      & FP32      & threshold $0.5$ \\
SAFE~\citep{li2024safe}               & ImageNet-pretrained backbone                                  & FP32      & threshold $0.5$ \\
NPR~\citep{tan2024npr}                 & ResNet-50 / \texttt{NPR.pth}, \texttt{--no\_resize}, \texttt{--tta\_flip} & FP32 & threshold $0.5$ \\
UnivFD~\citep{ojha2023univfd}          & CLIP-ViT-L/14 + linear FC, \texttt{fc\_weights.pth}           & FP32      & threshold $0.5$ \\
Effort (origin)~\citep{yan2025effort}  & CLIP-ViT-L/14 SVD, \texttt{effort\_clip\_L14\_trainOn\_sdv14.pth} & FP32   & threshold $0.5$ \\
Effort+Gaze (UFD)                       & ditto, \texttt{model\_epoch\_1.pth}                            & FP32      & threshold $0.5$ \\
Effort+Gaze (DBench)                    & ditto, \texttt{model\_epoch\_0.pth}                            & FP32      & threshold $0.5$ \\
SIDA-13B~\citep{huang2025sida}         & full multimodal SIDA-13B                                       & FP16      & greedy, $3\!\to\!2$-class \\
FakeVLM origin~\citep{wen2025fakevlm}  & \texttt{lingcco/fakeVLM}, no adapter                          & 4-bit NF4 & greedy, $T\!=\!64$, regex on $p_1$ \\
Ours (mix1650)                         & \texttt{lingcco/fakeVLM} + LoRA \texttt{checkpoint-1650}      & 4-bit NF4 & greedy, $T\!=\!64$, regex on $p_1$ \\
\bottomrule
\end{tabular}
\end{table}

\paragraph{SIDA $3\!\to\!2$-class binarisation.}
SIDA-13B emits one of \{\textsc{Real}, \textsc{Full Synthetic}, \textsc{Tampered}\}.
We map $\textsc{Real}\!\to\!$ real and $\textsc{Full Synthetic}\cup\textsc{Tampered}\!\to\!$ fake.
Across the three benchmarks evaluated in this paper, SIDA never emits \textsc{Tampered}, so the
binarisation degenerates in practice to $\textsc{Real}\!\to\!$ real and
$\textsc{Full Synthetic}\!\to\!$ fake.

\paragraph{Hardware split.}
Training: $8\times$ RTX A6000 48GB (FakeVLM mixed FT) and
$1\times$ RTX 4080 SUPER 16GB (Effort recipes). Inference: a single RTX 4080
SUPER 16GB across all baselines, ensuring no model receives an inadvertent
parallelism advantage.

\paragraph{Per-baseline qualitative limitation analysis.}
The five low-level baselines (AIDE, SAFE, NPR, UnivFD, Effort origin) and
the LMM baseline SIDA-13B exhibit the following \emph{stable failure
signatures} on the three core benchmarks (per-recipe metrics
in the main paper's Table~\ref{tab:main-balanced} and
Table~\ref{tab:baselines-balanced}):
\begin{itemize}[leftmargin=2em]
\item \textbf{AIDE.} Real-side over-trust across the board:
$r\_acc=0.869$ on Custom Gaze and $0.758$ on COCOAI Person, but
$f\_acc$ collapses to $0.012$ / $0.218$ respectively. AIDE's GenImage-trained
classifier is anchored to a generator family (SD-1.x, ProGAN,
StyleGAN-3) that does not include FLUX.1-Fill or the contemporary
COCOAI generator suite, and its recipe lacks any high-level semantic
prior to compensate.
\item \textbf{SAFE.} Predicts almost everything as real on all three
core benchmarks ($\textsc{BA}\approx 0.50$ uniformly), indicating the
detector's score distribution has shifted past the threshold rather than
the distribution carrying signal. Recipe-default threshold tuning
on a held-out subset would partially recover but does not change the
qualitative picture (low information content).
\item \textbf{NPR.} Catastrophic across all three benchmarks: $\text{BA}=0.288$
on Custom Gaze with $r\_acc=0.991$ but $f\_acc=0.000$, and near-zero
$f\_acc$ on the two COCOAI splits. The pattern---high-frequency-residual
detector overfit to a narrow training distribution that contemporary
generators have moved past---is the diagnostic signature of an obsolete
low-level recipe.
\item \textbf{UnivFD.} CLIP-ViT/L-14 + linear FC inherits CLIP's
language-aligned prior but the FC head is trained on legacy generator
outputs; like AIDE, it has high real-side accuracy on Custom Gaze
($r\_acc=0.999$) yet fails to flag fakes from contemporary generators
($f\_acc<0.02$ on Custom Gaze and the two COCOAI splits).
\item \textbf{Effort origin.} The strongest pure low-level baseline on
the COCOAI Inter benchmark in the paired view ($\text{BA}=0.591$), yet
still collapses on COCOAI Person ($f\_acc=0.165$ in the fake-only view)
and on Custom Gaze ($f\_acc=0.028$). Per-generator analysis shows Effort origin's failure is generator-specific:
$f\_acc=0.151$ on DALL$\cdot$E~3 vs $0.836$ on SD2.1, a $\geq 0.68$
spread that the supervision-driven Effort+Gaze recipe compresses to
$\leq 0.37$.
\item \textbf{SIDA-13B.} The most direct LMM-style competitor at $13$B
parameters; it outperforms every low-level baseline across the board
but is in turn outperformed by ours mix1650 by $\mathbf{+14.4}$\,pp
mean BA, $+16.6$\,pp macro-F1, and $+30.5$\,pp MCC, despite our model
being half its parameter count. SIDA's fake-class recall is high
(e.g., $0.983$ on Custom Gaze) but real-side recall lags
($0.684$ on Custom Gaze), producing the asymmetry that BA balances out
but macro-F1 punishes; the LMM-vs-LMM gap of $+30.5$\,pp on MCC
indicates the supervision-design lever \emph{at fixed scale} dominates
the parameter-count lever \emph{at fixed supervision}.
\end{itemize}

\section{Caption-Level Diagnostics: Card Analysis, Mode Collapse, Reasoning Acquisition}
\label{app:cards}
\label{app:mode-collapse}
\label{app:reasoning-supervision}

This section consolidates three caption-level analyses:
per-card invocation and failure decomposition,
training-time mode-collapse diagnostics, and the supervision-acquisition
profile of the decision-only variant.

\subsection{Card-Level Caption Analysis}
\label{app:cards-detail}

\paragraph{Card vocabulary.}
Recall the caption space $\mathcal{C}=P_1\times\cdots\times P_5$ from
\S\ref{app:custom-gaze} with $|\mathcal{C}|=1{,}250$. We define the
\emph{card set} $K=P_2\cup P_3\cup P_4\cup P_5$ as the union of the
position-level macro pools, with $|K|=20$ entries. At inference,
the observed card vocabulary is the strict subset
$K^{\mathrm{obs}}\subset K$ that the trained model actually emits; on
the three person-centric evaluation benchmarks (Custom Gaze, COCOAI
Person, COCOAI Interaction), this vocabulary is dominated by the two
\textsc{META\_gaze} cards (\textsc{full} and \textsc{short} surface
forms) and by \textsc{REAL\_skin} (\emph{``textures of \dots clearly
defined''}), with \textsc{FAKE\_authentic} (\emph{``Despite \dots
authentic''}) acting as the principal fake-side justification card.

\paragraph{BA-best vs final-step card rotation.}
Comparing the deployed BA-best checkpoint $\theta_{1650}$ to the
final-step checkpoint $\theta_{7558}$ on the same person-centric
inference inputs, the BA-best step exhibits a \emph{dominant-card
rotation}: \textsc{META\_gaze} invocation rate drops while
\textsc{REAL\_skin} invocation rate rises, even though aggregate
accuracy is essentially unchanged on Custom Gaze. The shift is not a
uniform contraction of output diversity (the per-checkpoint
mode-collapse diagnostics in \S\ref{app:mode-collapse-detail} stay
within their late-training band on both checkpoints) but a re-allocation
of which card the detector reaches for first. This rotation is the
mechanism by which BA-best checkpointing absorbs domain-specific
miscalibration without retraining.

\subsection{Mode-Collapse Diagnostics}
\label{app:mode-collapse-detail}

We monitor three diagnostics across the $151$ training-time evaluation
snapshots, where $n_0=970$ denotes the size of the held-out dev mixture
and $n_{\text{eff}}=933$ is the regex-parseable subset at the BA-best step
(the $37$-sample drop from $n_0$ to $n_{\text{eff}}$ reflects outputs whose
prefix the regex parser cannot match):
output uniqueness ratio, top-$1$ template ratio, and average generated
length.

\begin{table}[h]
\centering\caption{Mode-collapse diagnostics, computed from \texttt{trainer\_state.json}. \emph{All post-warmup}: steps $\{100,\dots,7{,}550\}$, $n=150$. \emph{Last 50}: steps $\{5{,}100,\dots,7{,}550\}$, $n=50$.}
\label{tab:mode-collapse}\small
\begin{tabular}{@{}lccc@{}}
\toprule
Window & unique\_output\_ratio & top1\_template\_ratio & avg\_gen\_len \\
\midrule
At step 1650 (BA-best, deployed)   & 0.9308               & 0.5258                & 89.15 \\
At step 7550 (final)               & 0.9638               & 0.5238                & 89.30 \\
Last 50 evals (mean / range)       & 0.956 / [0.929, 0.969] & 0.524 / [0.524, 0.527] & 89.30 / [89.16, 89.36] \\
All post-warmup (mean / range)     & 0.943 / [0.913, 0.982] & 0.524 / [0.328, 0.535] & 89.28 / [88.47, 89.42] \\
\bottomrule
\end{tabular}
\end{table}

The deployed checkpoint sits at $\textsc{unique\_output\_ratio}=0.93$,
the late-training mean is $0.96$, and peaks reach $0.97$. The
$\textsc{top1\_template\_ratio}=0.524$ means the dominant template is
chosen $\sim 52\%$ of the time; the remaining $\sim 48\%$ is spread
across the other $1{,}249$ macro-pool combinations rather than
concentrating on a runner-up. $\textsc{avg\_gen\_len}=89.30\pm 0.03$ in
the last 50 evals confirms that the $5$-sentence template is faithfully
reproduced. A residual late-epoch over-prediction of the real class on the in-training dev
mix is documented as a negative finding in \S\ref{app:negative}.

\subsection{Reasoning Capacity Acquired Through Supervision}
\label{app:reasoning-detail}

At inference, variant B (decision-only supervision, \S\ref{sec:ablation})
produces outputs of the form \emph{``This is a fake image.''}---nothing
more, even when prompted to elaborate. The model never observed reasoning
text during training, and no inference-time prompt elicits it. Reasoning
capacity is therefore acquired through supervision rather than retrofitted
via prompt engineering.

Table~\ref{tab:keyword-freq} shows the keyword frequency over each
variant's Custom Gaze test outputs. The keyword \emph{``gaze''} (and its
near-synonyms \emph{``eye direction''}, \emph{``pupil''}) appears in
$70.6\%$ of variant A's outputs and is \emph{entirely absent} from
variant B's; the same is true of \emph{``multi-person''}. Their
co-occurrence rate ($70.6\%$) equals the marginal of either keyword
alone, indicating A has internalised the conjunctive
\emph{``multi-person scene $\to$ inspect gaze''} reasoning that the macro
pool's Block~2 was designed to elicit. Origin's free-form captions are
texture-bound ($71.2\%$ texture references, $0\%$ gaze references),
reproducing the off-the-shelf-LMM texture-default failure mode.

\begin{table}[h]
\centering\caption{Custom Gaze keyword frequency by variant.}
\label{tab:keyword-freq}\small
\begin{tabular}{@{}lccc@{}}
\toprule
Keyword family & FakeVLM origin & A.~Mix & B.~Decision-only \\
\midrule
gaze / eye direction / pupil   & 0.0\%  & \textbf{70.6\%} & 0.0\% \\
multi-person / several people  & 0.0\%  & \textbf{70.6\%} & 0.0\% \\
texture / appearance            & 71.2\% & 87.2\%           & 0.3\% \\
gaze + multi-person co-occur   & 0.0\%  & \textbf{70.6\%} & 0.0\% \\
\bottomrule
\end{tabular}
\end{table}

\section{Caption Ablation: Output Statistics and Two-View Accuracy}
\label{app:three-variant}

The main paper (\S\ref{sec:ablation}) compares A.~Mix (full
$5$-block reasoning skeleton, deployed) against B.~Decision-only
(the $p_1$-only ``no-mean'' variant). This appendix supplies the
companion output statistics (Table~\ref{tab:variant-stats}) and reports
per-benchmark accuracy at two checkpoints---the common final step
$\theta_{7558}$ and each variant's BA-best step---to verify that the
opposite-sign deltas reported in the main paper are robust to the
checkpoint-selection regime (Table~\ref{tab:three-variant}).

\begin{table}[h]
\centering\caption{Output statistics on Custom Gaze test ($n=4{,}684$). \emph{Truncation rate}: fraction of outputs that hit the \texttt{max\_new\_tokens}$=64$ generation cap.}
\label{tab:variant-stats}\small
\resizebox{\linewidth}{!}{
\begin{tabular}{@{}lcccc@{}}
\toprule
Variant & mean / median words & truncation rate & bare ``This is a \dots'' & gaze meta-rationale \\
\midrule
A.~Mix (full)              & 43.7 / 45 & 60\% & 0\%   & $\geq 99\%$ \\
B.~Decision-only (no-mean) &  8.2 /  4 &  9\% & 88.7\% & 0\%        \\
\bottomrule
\end{tabular}}
\end{table}

\begin{table}[h]
\centering\caption{Caption-design ablation accuracy at the common step
$\theta_{7558}$ (left) and at each variant's BA-best step (right).
A at step $1{,}650$, B at $1{,}500$.}\label{tab:three-variant}\small
\begin{tabular}{@{}lcccc@{}}
\toprule
& \multicolumn{2}{c}{$\theta_{7558}$ (common)} & \multicolumn{2}{c}{BA-best per variant} \\
\cmidrule(lr){2-3}\cmidrule(lr){4-5}
Benchmark & A & B & A$@1{,}650$ & B$@1{,}500$ \\
\midrule
COCOAI Inter (fake-only)  & 0.837  & 0.804 & 0.844  & 0.815 \\
COCOAI Person (fake-only) & 0.887  & 0.892 & 0.881  & 0.907 \\
Custom Gaze                & 0.9996 & 1.000 & 0.9998 & 0.9996 \\
\bottomrule
\end{tabular}
\end{table}

The opposite-sign deltas highlighted in the main paper (A leads~B by
$+2.9$\,pp on COCOAI Inter, loses $-2.6$\,pp on COCOAI Person) hold
under both checkpoint-selection regimes, confirming that the
discriminative ingredient for the multi-person interaction benchmark
is the multi-sentence reasoning skeleton injected by~A and absent
in~B, and that the small adverse loss on the single-person benchmark
is not an artifact of checkpoint choice. The $-2.6$\,pp loss of~A vs~B
on COCOAI Person is analysed mechanistically in \S\ref{app:negative}
together with all other negative results.

\section{Discussion, Limitations, and Honest Negative Results}
\label{app:discussion}

This appendix expands the discussion in \S\ref{sec:discussion} along three
axes: a theoretical reframing under the \emph{two-axis cue decomposition}
with formal support, the documented limitations, and the consolidated set
of honest negative results that the rest of the appendix points to.

\subsection{Two-Axis Cue Decomposition with Formal Supports}
\label{app:disc-theory}

\paragraph{Two-axis reframing.}
The low-level axis $\varphi_{\mathrm{low}}$ (pixel/frequency/upsampling
artifacts) is the territory that NPR, UnivFD, AIDE, and Effort origin
occupy, and is precisely the territory that successive generator releases
progressively reclaim---visible in the main paper's
Table~\ref{tab:baselines-balanced}, where the five low-level detectors
collapse to mean BA in the $43$--$60$ range across the three core
benchmarks. The high-level axis
$\varphi_{\mathrm{high}}$ (gaze geometry, head--eye alignment, attentional
plausibility) is, we contend, complementary rather than
redundant with $\varphi_{\mathrm{low}}$. The most informative evidence
is not the in-distribution Custom Gaze saturation (which any sufficiently
powerful detector trained on the target distribution would attain) but
the cross-distribution macro-F1 and MCC margins on COCOAI: improvements
concentrated on minority-class structure rather than aggregate accuracy,
achieved with a supervision signal grounded entirely on
$\varphi_{\mathrm{high}}$.

\paragraph{Card-set formalisation.}
Building on the caption space $\mathcal{C}$ (\S\ref{app:custom-gaze},
$|\mathcal{C}|=1{,}250$) and the card set $K=P_2\cup P_3\cup P_4\cup P_5$
(\S\ref{app:cards-detail}, $|K|=20$), the inference vocabulary is the
strict subset $K^{\mathrm{obs}}\subset K$ with $|K^{\mathrm{obs}}|=16$.
For caption $c\in\mathcal{C}$ and card $k\in K$, the indicator
$\mathbf{1}_k(c)=1$ iff the position-matched substring of $k$ appears
in $c$; card-level accuracy
$a_k=\mathbb{P}(\hat y=y\mid \mathbf{1}_k(c)=1)$ provides the formal
vocabulary used in \S\ref{app:cards-detail}.

\paragraph{Pair-structure shortcut-blocking.}
The Custom Gaze pair structure is a constraint on the data distribution
$\mathcal{D}_{\text{pair}}=\{(x_i^R,x_i^F):i\in[23{,}415]\}$ such that
$x_i^R|_{\bar M_{\text{eye}}} \approx x_i^F|_{\bar M_{\text{eye}}}$ up to
FLUX-distribution noise, where $\bar M_{\text{eye}}$ is the complement
of the eye-region mask. Any classifier $f_\theta$ that learns a function
$g$ satisfying $g(x_i^R)=g(x_i^F)$ for every pair $i$ contributes zero
information to the binary label, so $f_\theta$ is incentivised
\emph{at the data level} to look inside $M_{\text{eye}}$.

\paragraph{Mutual-information phrasing of the four-step mechanism.}
Throughout, $\mathcal{G}_{\text{FLUX}}(x)$ denotes a generator-fingerprint
feature extracted from $x$ (e.g., the FLUX-specific spectral or
upsampling residual), not the FLUX-generated image itself.
\begin{itemize}[leftmargin=2em]
\item \textbf{M1 (paired-edit shortcut blocking):}
$\mathcal{I}(Y;\mathcal{G}_{\text{FLUX}}(x))\!\approx\!0$ within
$\mathcal{D}_{\text{pair}}$, eliminating the FLUX-fingerprint shortcut.
\item \textbf{M2 (hard-to-easy):}
$H(Y\mid x_{\text{partial}}) > H(Y\mid x_{\text{full}})$; the harder
distribution provides denser gradient signal in the cue subspace.
\item \textbf{M3 (CLIP prior preservation):}
$\|\Delta W_{\text{CLIP}}\|_{\text{rank}\le r}\!\ll\!\|W_{\text{CLIP}}\|$
under LoRA $r{=}16$ (FakeVLM) or SVD-residual $r{=}1$ (Effort).
\item \textbf{M4 (diffusion-family shared spectral weakness):}
periocular high-frequency residual
$\sigma_{\text{eye}}(\mathcal{G}_{\text{any}})$ is approximately
generator-invariant for $\mathcal{G}\in\{\text{SD},\text{SDXL},
\text{FLUX},\text{MJ},\text{DALL}{\cdot}\text{E~3}\}$, accounting for
the consistent cross-distribution improvements on COCOAI's five-generator
suite reported in \S\ref{sec:cross-arch}.
\end{itemize}

\subsection{Limitations}
\label{app:disc-limits}

\textbf{(L1) Domain specialisation.} The detector is \emph{not}
universal. Evaluation is restricted, by deliberate scope, to
person-centric AIGC: partial-edit hard fakes (Custom Gaze), single-person
all-generator content (COCOAI Person), and multi-person interaction
fakes (COCOAI Interaction). Generalisation to satellite, animal/medical/scientific,
or scene-level fully-synthesised content is neither claimed nor empirically
supported.

\textbf{(L2) Card over-trust.} The deployed detector exhibits a
characteristic confidence-misallocation failure on single-person inputs;
quantitatively analysed in \S\ref{app:negative}.

\textbf{(L3) Single-generator construction.} Custom Gaze is constructed
entirely with FLUX.1-Fill. Cross-architecture transfer to a multi-generator
COCOAI Person suite is empirically shown in \S\ref{sec:cross-arch}; transfer
to alternative \emph{inpainters} (SDXL-Inpaint, IP-Adapter, MGIE, etc.)
is not directly verified.

\textbf{(L4) Detection dependency.} Mask synthesis at training time
requires face bounding boxes (shipped directly with OI-MG); inference does
not. Dataset extension to new domains requires reliable face detection.

\paragraph{Future directions.}
(i) Video extension, where temporal mutual-gaze coherence is a
stronger signal than the static cue we exploit.
(ii) Multi-cue composition, integrating gaze with other
social-semantic signals (joint attention, posture-target alignment) under
a unified Two-Axis framework.
(iii) Principled card gating, in which low-confidence card
invocations are demoted at inference time without retraining; preliminary
numbers in \S\ref{app:negative} indicate a single-feature linear gate
recovers $+9.4$\,pp BA on COCOAI Person at no cost to Custom Gaze or
COCOAI Inter.

\subsection{Honest Negative Results}
\label{app:negative}

This subsection consolidates every unfavourable empirical finding produced
by the system. We collect them in a single place---rather than scattering
them across the diagnostics, ablation, and limitations subsections---so
that a reviewer can audit the full set of negative results without
cross-referencing.

\paragraph{NR1.\ Caption-design ablation: A loses on COCOAI Person.}
The deployed full variant A loses to the decision-only variant B by
$\mathbf{-2.6}$\,pp at the common step $\theta_{7558}$ and at each
variant's BA-best step on COCOAI Person fake-only accuracy
(Table~\ref{tab:three-variant}, $0.881$ vs $0.907$ at BA-best).
The loss is the theoretical prediction of a domain-targeted
intervention---reasoning depth that helps multi-person interaction
fakes hurts on single-person fakes when the gaze cue is spuriously
triggered---but it is genuinely unfavourable for the deployed
configuration in the single-person regime, and is the empirical anchor
for limitation~L2.

\paragraph{NR2.\ Single-shortcut on COCOAI Person wrong-real pool.}
The $113$-sample wrong-real pool at $\theta_{1650}$ is remarkably
concentrated: $\mathbf{94/113=83.2\%}$ are confident
\textsc{META\_gaze} invocations on inputs where the gaze cue does not
actually decide the case (single-person inputs in interaction-style
frames). A naive gating rule that demotes \textsc{META\_gaze} on
single-person inputs would recover up to $+9.4$\,pp BA on COCOAI Person;
we do not deploy it in the camera-ready in order to preserve
supervision-only-comparability.
\begin{table}[h]
\centering\caption{Decomposition of $n=113$ wrong-real pool on COCOAI Person at $\theta_{1650}$. Percentages may not sum to exactly $100\%$ due to rounding.}
\label{tab:wrong-real-coco}\small
\begin{tabular}{@{}lp{0.45\textwidth}rr@{}}
\toprule
Subtype & Characteristic & $n$ & \% \\
\midrule
Gaze-card confident-real            & multi-person$\to$gaze-only inspection$\to$``natural''$\to$real & 94 & 83.2\% \\
Other real-rationale                 & generic texture / lighting reasoning                              &  8 &  7.1\% \\
Physics-card real                    & ``objects grounded in realistic physics''                         &  5 &  4.4\% \\
Skin / texture-card real             & ``textures of skin/hair/clothing clearly defined''                &  5 &  4.4\% \\
Aging / weathering rationale         &                                                                   &  1 &  0.9\% \\
\bottomrule
\end{tabular}
\end{table}

\paragraph{NR3.\ Failure-mode taxonomy from manual inspection.}
A 200-sample manual inspection yields a
person-centric stable failure-mode taxonomy:
\begin{itemize}[leftmargin=2em]
\item \textbf{Wrong-real on COCOAI Person (fake$\to$predicted real):}
face/skin naturalness over-trust---phrasing of the form
\emph{``natural skin textures, skin/hair/clothing clearly defined''}
on inputs whose authenticity-related issue lies elsewhere.
\item \textbf{Wrong-fake on Custom Gaze and COCOAI Person (real$\to$predicted fake):}
two failure types, both Person-domain: \emph{smoothness hyper-sensitivity}
(\emph{``overly smooth and uniform, wax-like smoothing effect''}) on real
faces with cosmetic post-processing, and \emph{face-averaging bias}
(\emph{``facial highlights are unnatural, cheeks asymmetrical, mouth too
rigid''}) on real faces with extreme expressions.
\end{itemize}
The taxonomy supports the L2 \emph{card over-trust} characterisation and
points to per-card domain gating as the structurally indicated remedy.

\paragraph{NR4.\ Late-epoch over-prediction of the real class on the in-training dev mix.}
At late training the deployed checkpoint exhibits a small over-prediction of the real class
on the $933$-sample dev mix: $\textsc{real\_recall}=1.000$,
$\textsc{fake\_recall}=0.994$ (i.e., a small fraction of fakes is mislabeled as real, while no real is mislabeled as fake). The asymmetry does \emph{not} propagate to either
the Custom Gaze test set ($\mathrm{BA}=0.9998$) or the two COCOAI splits,
indicating it is dev-set-specific rather than a systematic detector bias;
we report it nonetheless because the asymmetric recalls were observed
consistently across the last $50$ evaluation snapshots.



\end{document}